\pdfoutput=1

\documentclass[11pt]{article}

\usepackage[final]{acl}

\usepackage{times}
\usepackage{latexsym}

\usepackage[T1]{fontenc}

\usepackage[utf8]{inputenc}

\usepackage{microtype}

\usepackage{inconsolata}

\usepackage{graphicx}

\usepackage[utf8]{inputenc}

\usepackage{amsmath}
\usepackage{amsfonts}
\usepackage{amsthm}
\usepackage{bm}

\usepackage{float}

\def\eqref#1{equation~\ref{#1}}

\def\1{\bm{1}}

\def\vb{{\bm{b}}}

\def\vx{{\bm{x}}}

\def\mW{{\bm{W}}}
\def\mX{{\bm{X}}}

\DeclareMathAlphabet{\mathsfit}{\encodingdefault}{\sfdefault}{m}{sl}
\SetMathAlphabet{\mathsfit}{bold}{\encodingdefault}{\sfdefault}{bx}{n}

\usepackage{ amssymb }
\usepackage{pifont}
\usepackage{subcaption}
\newcommand{\Loss}{\mathcal{L}}
\newcommand{\cmark}{\ding{51}}%
\newcommand{\xmark}{\ding{55}}%

\title{Efficient and Interpretable Grammatical Error Correction \\
with Mixture of Experts}

\author{Muhammad Reza Qorib\footnotemark[2]{}, Alham Fikri Aji\footnotemark[3]{}\and Hwee Tou Ng\footnotemark[2]{} \\
        { }\footnotemark[2]{}Department of Computer Science, National University of Singapore \\
        { }\footnotemark[3]{}NLP department, MBZUAI \\
        \texttt{mrqorib@u.nus.edu}, \   \texttt{alham.fikri@mbzuai.ac.ae}, \
        \texttt{nght@comp.nus.edu.sg}}

\begin{document}
\maketitle
\begin{abstract}
Error type information has been widely used to improve the performance of grammatical error correction (GEC) models, whether for generating corrections, re-ranking them, or combining GEC models. Combining GEC models that have complementary strengths in correcting different error types is very effective in producing better corrections. However, system combination incurs a high computational cost due to the need to run inference on the base systems before running the combination method itself. Therefore, it would be more efficient to have a single model with multiple sub-networks that specialize in correcting different error types. In this paper, we propose a mixture-of-experts model, MoECE, for grammatical error correction. Our model successfully achieves the performance of T5-XL with three times fewer effective parameters. Additionally, our model produces interpretable corrections by also identifying the error type during inference.\footnote{The source code and models can be accessed at \url{https://github.com/nusnlp/moece}.}
\end{abstract}

\begin{table*}[bht]
\centering
\begin{tabular}{p{0.1\linewidth} p{0.85\linewidth}}
\hline
Source & The rich people will buy a car but the poor people always need to use a bus or taxi . \\
Correction & Rich people will buy a car , but poor people always need to use a bus or taxi .\\
Edits & (0, 2, 'Rich', \texttt{DET}), (7, 7, ',', \texttt{PUNCT}), (8, 9, '', \texttt{DET}) \\
\hline
\end{tabular}
\caption{\label{tab:example}
Example of a GEC source sentence, its correction, and the edits. Each edit is represented by the start index, the end index, the replacement string, and the error type. \texttt{DET} denotes a determiner error while \texttt{PUNCT} denotes a punctuation error.
}
\end{table*}

\begin{figure*}[bht]
\centering
\includegraphics[width=0.9\linewidth]{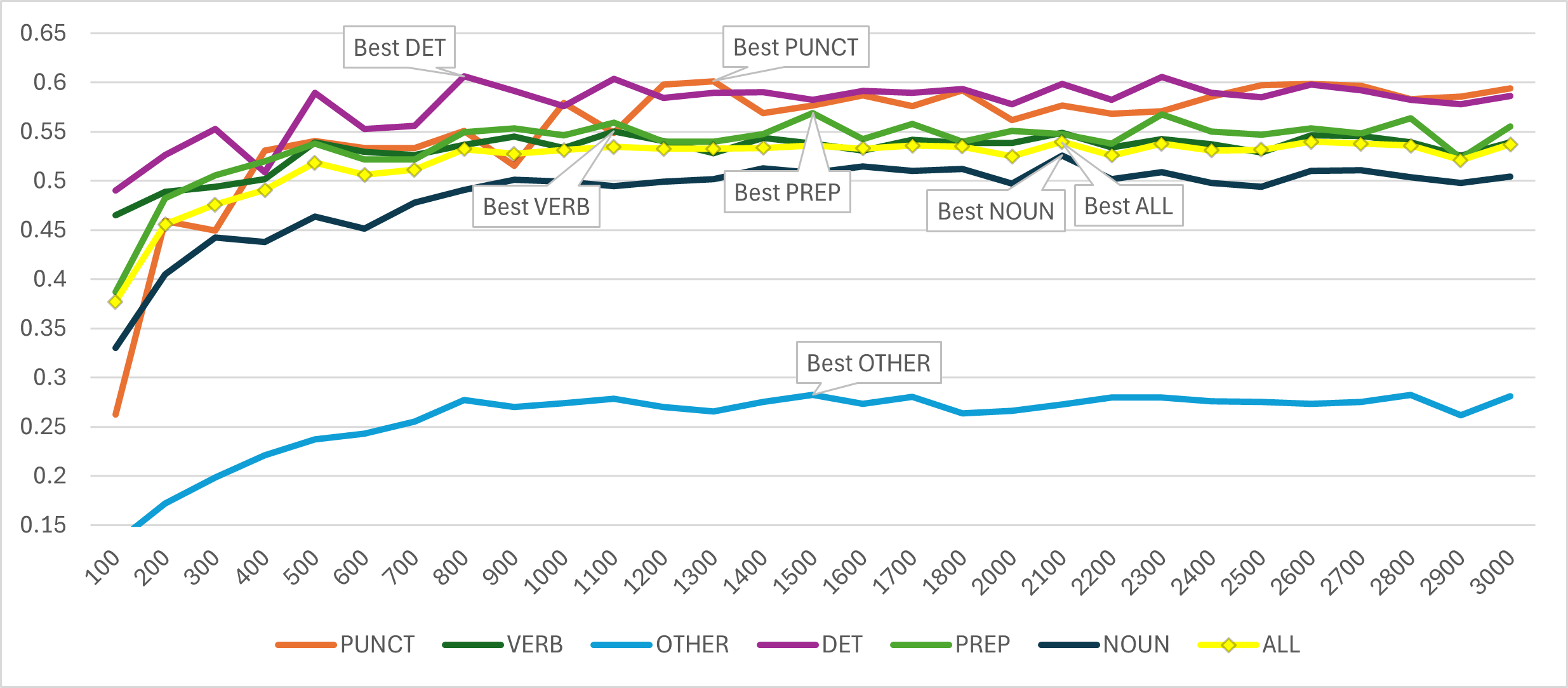}
\caption{$F_{0.5}$ scores of a T5-v1.1-Base model on the six most frequent error types and all error types (\texttt{ALL}) in the BEA-2019 development set at different numbers of training steps.}\label{fig:t5-base}
\end{figure*}

\section{Introduction}
\label{sec:intro}
Grammatical error correction (GEC) is a task that aims to detect and correct any errors in a given text. Through scientific progress over the past decades, grammatical error correction has outgrown its name. GEC not only deals with grammatical errors but also includes the correction of misspellings, orthographic errors, semantic errors, and more \cite{gec_survey}.

A GEC model receives a possibly erroneous text and should produce a corrected version of the text with minimal modification. An example is given in Table~\ref{tab:example}. In the early days of GEC, people approached this problem by building a specific classifier for each error type, which was later replaced by data-driven approaches adopted from the machine translation task \cite{chollampatt-etal-2016-neural}.

Even after adopting end-to-end approaches using neural network models, the error type information remains useful. Some recent models approach the problem by separating the error detection and correction models \cite{yuan-etal-2019-neural_short, li-etal-2023-templategec_short}, training a single model for both detection and correction \cite{omelianchuk-etal-2020-gector_short}, or training a model to predict the edit/transformation tag, which can include the error type \cite{stahlberg-kumar-2020-seq2edits_short}.

Error type information is useful not only for building the grammar correction model but also for reranking the outputs of GEC models \cite{sorokin-2022-improved_short, 10.5555/3504035.3504741} and combining outputs of GEC models \cite{kantor-etal-2019-learning,gec-ip,qorib-ng-2023-system_short}. GEC system combination has been very effective in improving the state of the art, and it has been repeatedly demonstrated that it works best when the base systems have complementary strengths \cite{susanto-etal-2014-system_short,ddc}, such as which error types they can correct more accurately. However, system combination is computationally expensive because it needs to run inference on each base system first before running the combination method itself.

It would be desirable to just have one model with multiple sub-networks that specialize in different aspects, such as which error types it can correct more accurately. The capability of a neural network to correct one error type is not necessarily transferable to correcting other error types, as \citet{qorib-etal-2022-frustratingly_short} report that state-of-the-art GEC models are often less accurate in correcting certain error types than weaker GEC models. This could be caused by task interference during the training of the model. When fine-tuning T5-v1.1-Base for grammatical error correction, we observe that the model becomes less accurate at correcting punctuation (\texttt{PUNCT}) and preposition (\texttt{PREP}) error types when it has the best overall (\texttt{ALL}) performance (Figure~\ref{fig:t5-base}). We also notice that the model's accuracy in correcting preposition errors decreases when its accuracy in correcting punctuation errors increases (e.g., training step 1600 to 2000 in Figure~\ref{fig:t5-base}).

As such, a neural network that has separate sets of parameters that specialize in correcting different error types could be beneficial in modeling grammatical error correction. Such a model is computationally more efficient as it does not need to go through multiple threads of inference necessary in system combination. For that purpose, we propose a mixture-of-experts model for grammatical error correction. To the best of our knowledge, we are the first to do so. This study focuses on building an MoE model for GEC through transfer learning from a non-MoE (dense) sequence-to-sequence model.

\begin{figure*}[thb]
\centering
\includegraphics[width=0.9\linewidth]
{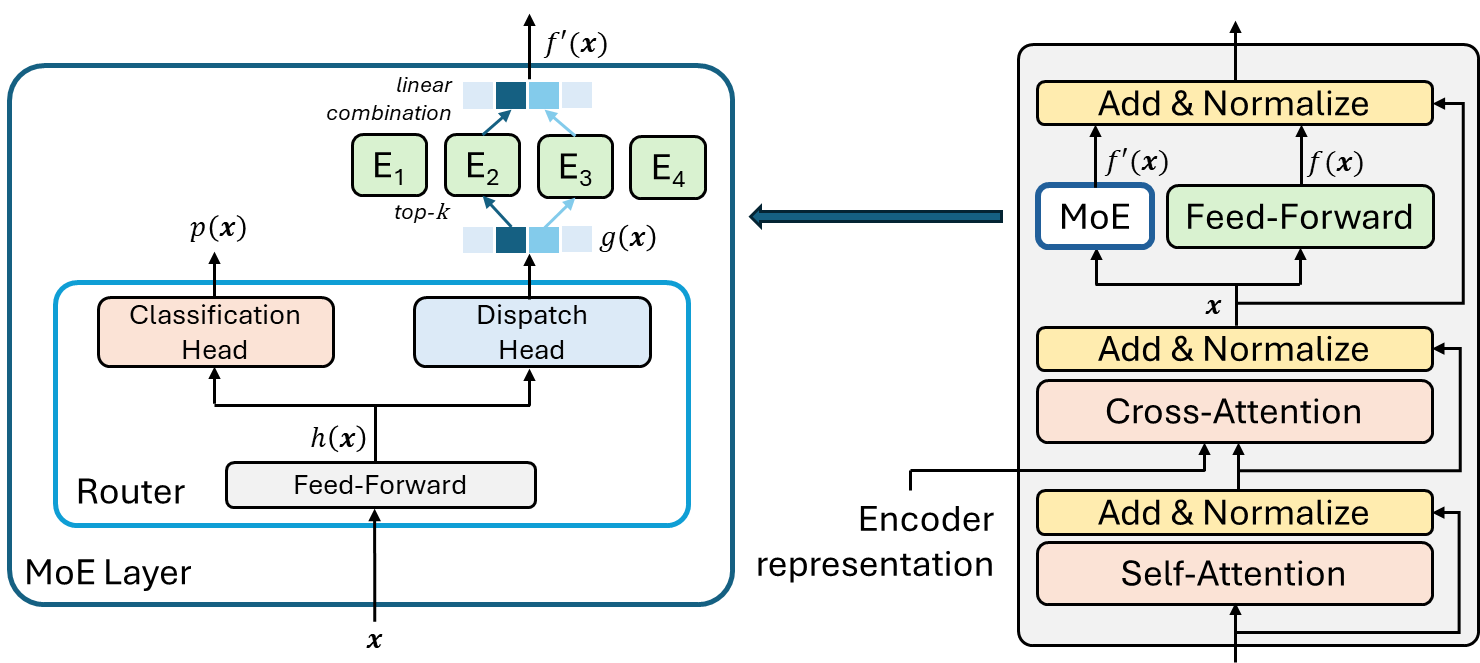}
\caption{Illustration of a transformer block with an MoE layer with $M=4$ and $K=2$. To simplify the notation, $\vx$ represents the input of the MoE layer and the feed-forward layer instead of the transformer block.}
\label{fig:architecture}
\vspace*{-0.5\baselineskip}
\end{figure*}

\section{Mixture of Experts}
\label{sec:moe}
Mixture of experts (MoE) is a learning procedure for a system that can be decomposed into multiple separate networks, with each network representing an expert at a particular task \cite{Jacobs1991AdaptiveMO}. Mixture of experts has been reported to perform well on tasks that can be broken down into different sub-tasks, such as multilingual machine translation \cite{kudugunta-etal-2021-beyond-distillation_short}, and single tasks that have multiple objectives, such as video recommendation \cite{li-etal-video_short}.

When used with the transformer architecture, MoE is typically applied by replacing the feed-forward layer $f(\vx)$ in a transformer block with an MoE layer\footnote{The notation of $f'(\vx)$ represents a function that replaces $f(\vx)$, rather than the derivative of $f(\vx)$.} $f'(\vx)$. MoE is applied independently to some or all transformer blocks in the network. An MoE layer consists of a set of experts $\{E_1, E_2, \dots, E_M\}$ with parameters $\phi_i, i\in \{1, \dots, M\}$, a router or a gating function $r$ that selects a subset of $K$ experts, and optionally an aggregation function $g_{i}$ that produces the weight for expert $i$. The aggregation function can also be the same as the router.
Let $\vx$ denote the vector representation of a token from the previous layer. We formalize the MoE layer in Equation \ref{eq:moe} below:

\begin{equation}
f'(\vx) = \sum_{k=1}^{K}\frac{g_{k}(\vx)}{\sum_{j=1}^{K} g_{j}(\vx)}E_{k}(\vx;\boldsymbol{\phi_k})
\label{eq:moe}
\end{equation}

One problem that may arise in MoE is one expert getting chosen more often than others, making it receive more gradient updates and subsequently making it more preferred by the router, a self-reinforcing problem. To avoid this issue, the number of tokens in a sequence $\mX = \{\vx_1, \vx_2, \ldots, \vx_N\}$ that an expert can process is often limited to a threshold called expert capacity, and a load balancing loss is used to encourage a more balanced expert allocation. If an expert already reaches the maximum capacity, subsequent tokens that are assigned to that expert do not go through the expert layer. The load balancing losses are calculated based on the fraction of the input that goes into expert $i$, represented by $w_i$.

\subsection{GShard}
\citet{lepikhin2021gshard} propose to penalize the model based on the mean square of $w_i$, but the top-k operation to select the experts is not differentiable. As such, it approximates the loss function by multiplying the average routing score for expert $i$ on all tokens ($v_i$) with the fraction of tokens that are allocated to it ($w_i$), then takes the average of that score from all experts. The loss function is formalized in Equation \ref{eq:gshard} below.
\begin{align}
v_i &= \frac{1}{N}\sum_{\vx \in \mX}g_i(\vx) \nonumber \\
\Loss_b &= \frac{1}{M}\sum_{i=1}^M (w_i \times v_i) \label{eq:gshard}
\end{align} 

In GShard, the router chooses the top two experts ($E_{i_1}$, $E_{i_2}$) for each token. However, the second expert will be ignored if its routing score is too small. The router sends the token $\vx$ to the second expert with a probability proportional to the routing score $g_{i_2}(\vx)$.

\subsection{SwitchTransformer}
SwitchTransformer \cite{10.5555/3586589.3586709} follows the balancing loss of GShard, but instead of taking the average, they take the sum and then multiply it again with the number of experts, as shown in Equation \ref{eq:switch} below.
\begin{align}
\Loss_b &= M \times \sum_{i=1}^M (w_i \times v_i) \label{eq:switch}
\end{align} 

The reason is to keep the loss constant and invariant to the number of experts. In the case of uniform routing, $w_i=\frac{1}{M}$ and $g_i=\frac{1}{M}$, so $\Loss_b = 1$.

\citet{10.5555/3586589.3586709} argue that sending each token to only one expert is sufficient. This challenges the conjecture of \citet{shazeer2017} that sending a token to more than one expert is necessary. They report that choosing one expert for each token preserves model quality, reduces routing computation, and performs better.

\section{Method}
In this section, we explain how we build MoECE (\textbf{M}ixture \textbf{o}f \textbf{E}rror \textbf{C}orrection \textbf{E}xperts), a mixture-of-experts model for grammatical error correction.

\subsection{MoE Layer}
\label{sec:moe_layer}
We follow the common approach of building MoE explained in Section \ref{sec:moe} with one modification. Instead of replacing the feed-forward layer with an MoE layer, we augment the feed-forward layer with an MoE layer (Figure~\ref{fig:architecture}). The output of the $i$-th transformer block in a standard transformer is the normalized $\vx_i + f_i(\vx_i)$ and the output of the transformer block in common MoE approaches is the normalized $\vx_i + f_i'(\vx_i)$. In our model, the output of the transformer block with an MoE layer is the normalized $\vx_i + f_i(\vx_i) + f_i'(\vx_i)$. While this design increases the computation cost during training, no additional cost is incurred during inference by merging the weights of the feed-forward layer with the experts.

\subsection{Router}

We hypothesize that an optimal expert allocation in a mixture of experts for grammatical error correction involves having each expert focus on correcting particular error types. With this goal in mind, the router needs to know what error type the current token has, so that the router can direct it to the appropriate expert. We approach this by adding a classification head to the router and training it with an auxiliary loss. As such, for every input token, the router predicts the error type through the classification head and determines the expert allocation through the dispatch head. 

We design the router in each MoE layer to be a 2-layer neural network with two outputs (Figure~\ref{fig:architecture}). The input dimension of the gate is the same as the model dimension of the transformer block. The output dimension of the classification head is the number of possible error types $|T|$, while the output dimension of the dispatch head is the number of experts $|M|$. Let $\vx$ denote an input token from the previous layer, and let $\sigma$ denote the softmax function. We formally describe the routing function in Equation 
(\ref{eq:router_class} -- \ref{eq:router_dispatch}) below.
\begin{align}
h(\vx) &= \mW_h \times \vx + \vb_h \nonumber \\
p(\vx) &= \sigma(\mW_p \times h(\vx) + \vb_p) \label{eq:router_class}\\
g(\vx) &= \sigma(\mW_g \times h(\vx) + \vb_g) \label{eq:router_dispatch}
\end{align}

While the classification head guides the hidden representation of the router $h$ to route the token based on the error type, the dispatch head may unsatisfactorily route tokens with different error types to a single expert. To avoid that, we employ the expert capacity limitation and load balancing loss of GShard and SwitchTransformer. Similar to previous work, we also use the routing function $g$ as the aggregation function.

\subsection{Training Objective}
The model is trained to minimize the combination of three loss functions, the cross-entropy loss $\Loss_c$ on the prediction of the corrected text, the cross-entropy loss of the error type $\Loss_e$ from the router, and the load balancing loss $\Loss_b$ from the router. The contribution of the error type loss is controlled by a hyper-parameter $\alpha$ while the contribution of the load balancing loss is controlled by another hyper-parameter $\beta$. Let $L$ denote the number of MoE layers. The final loss function is given below.
\begin{align}
\Loss &= \Loss_c + \alpha \times \frac{1}{L} \sum_{l=1}^L \Loss_{e_l} + \beta \times \frac{1}{L} \sum_{l=1}^L\Loss_{b_l} \label{eq:final_loss}
\end{align}

\section{Experiments}
We transformed the pre-trained T5-v1.1 \cite{JMLR:t5} language models into mixture-of-experts models. We trained two sizes of the model, a base model based on T5-v1.1-Base and a large model based on T5-v1.1-Large. For each model size, we trained two models with different routers, one with the GShard router (MoECE-GS) and another with the SwitchTransformer router (MoECE-ST). We apply an MoE layer with 7 experts to all transformer blocks in the decoder except the first block and share the parameters of the routers in all transformer blocks.

\begin{table}[htb]
\centering
\begin{tabular}{l | l | r r }
\hline
\textbf{Name} & \textbf{Type} & \textbf{\# sent} & \textbf{\# ref}\\
\hline
cLang-8 & Train & 2,372,119 & 1 \\
\hline
BEA-2019 & Dev & 4,384 & 1\\
\hline
CoNLL-2014 & Test & 1,312 & 2 \\
BEA-2019 & Test & 4,477 & 5 \\
CWEB-G & Test & 3,981 & 2 \\
CWEB-S & Test & 2,864 & 2 \\
\hline
\end{tabular}
\caption{\label{tab:data}
GEC datasets used in our experiments. \# sent refers to the number of sentences while \# ref refers to the number of reference annotations. Dev refers to the BEA-2019 development set.
}
\vspace*{-0.5\baselineskip}
\end{table}

\begin{table*}[thb]
\centering
\begin{tabular}{l | r | r | r r r | r r r }
\hline
{} & {} & \textbf{BEA-2019} & \multicolumn{3}{c|}{\textbf{CoNLL-2014 Test}} & \multicolumn{3}{c}{\textbf{BEA-2019 Test}}\\
\textbf{Model} & \textbf{EPC} & \textbf{Dev (F\textsubscript{0.5})} & \textbf{P} & \textbf{R} & \textbf{F\textsubscript{0.5}} & \textbf{P} & \textbf{R} & \textbf{F\textsubscript{0.5}}\\
\hline
T5-v1.1-Base & 248M & 53.97 & 72.43 & 48.38 & 65.88 & 72.82 & 62.11 & 70.39\\
MoECE-GS-Base & 282M & \textbf{55.28} & 73.00 & 48.53 & 66.31 & 74.59 & 62.11 & \textbf{71.71} \\
MoECE-ST-Base & 248M & 54.84 & 72.75 & 49.31 & \textbf{66.43} & 73.99 & 62.00 & 71.24 \\
\hline
T5-v1.1-Large & 783M & 55.85 & 73.18 & 49.46 & 66.78 & 75.65 & 64.59 & 73.15 \\
MoECE-GS-Large & 917M & 56.42 & 74.29 & 50.21 & \textbf{67.79} & 76.91 & 64.54 & \textbf{74.07} \\
MoECE-ST-Large & 784M & \textbf{56.68} & 73.60 & 51.25 & 67.69 & 75.95 & 65.78 & 73.67 \\
\hline
\end{tabular}
\caption{\label{tab:bea_conll}
MoECE performs better than the comparable dense model on the BEA-2019 development set, CoNLL-2014 test set, and BEA-2019 test set. EPC is the effective parameter count.
}
\vspace*{-0.5\baselineskip}
\end{table*}

\subsection{Implementation}
We implement our T5 model using the modification of the fairseq framework \cite{ott2019fairseq} by Applica AI\footnote{\url{https://github.com/applicaai/fairseq/tree/applica-t5}} and augment it with the MoE implementation from Fastmoe \cite{he2021fastmoe}. In Fastmoe, the GShard loss function is multiplied by the square of the number of experts ($M^2$) to make the loss magnitude not affected by the number of experts in the layer. We follow this implementation to standardize the experiments. As such, the difference between the GShard and SwitchTransformer variants in our model is in the number of experts selected per token and when the loss is calculated. In GShard, the loss is calculated before limiting the expert by its capacity, while in SwitchTransformer it is calculated after.

\subsection{Datasets and Evaluation}
\label{sec:data_and_eval}
Following \citet{rothe-etal-2021-simple_short}, we train the model with the cLang-8 dataset. The cLang-8 corpus is a GEC dataset that was made by relabeling the raw version of the Lang-8 dataset with the output of a big multilingual GEC model. Training a GEC model with cLang-8 is reported to produce a better GEC model than training it with the original Lang-8 dataset \cite{sorokin-2022-improved_short}. We augment the training data with the error types produced by ERRANT \cite{errant}. During the development of the model, we use the BEA-2019 development set \cite{bryant-etal-2019-bea_short} as the validation data to choose the hyper-parameters.

We evaluate the model on standard GEC benchmarks, the CoNLL-2014 test set and the BEA-2019 test set. The model is evaluated on the $F_{0.5}$ score produced by the M2Scorer \cite{dahlmeier-ng-2012-better_short} for the CoNLL-2014 test set and the $F_{0.5}$ score produced by the official blind scorer\footnote{\url{https://codalab.lisn.upsaclay.fr/competitions/4057}} for the BEA-2019 test set. In addition to the standard benchmarks, we also evaluate our model on out-of-domain datasets, which are CWEB-G and CWEB-S \cite{flachs-etal-2020-grammatical_short}. Most of GEC training and test data were made from student essays in an academic setting, but CWEB datasets were made from texts from various websites on the Internet. The CWEB-G test set was made from generic websites while the CWEB-S test set was made from websites of official institutions, such as governments, schools, and museums. We report the statistics of the datasets in Table~\ref{tab:data}. We perform a bootstrap resampling test on 100 samples of the models' outputs to measure statistical significance.

We compare our model to baselines and previous work that have comparable effective parameter count (EPC) \cite{Fedus2022Review}. The effective parameter count only considers the number of parameters that are active or used in a single forward pass. Since in MoE only a few experts are used at a time to generate the corrections, the unused experts do not contribute to the computational cost. Therefore, it is more appropriate to compare an MoE model based on the effective parameter count rather than the total number of parameters.

\begin{table}[thb]
\centering
\begin{tabular}{l | r r }
\hline
\textbf{Model} & \textbf{CWEB-G} & \textbf{CWEB-S}\\
\hline
T5-v1.1-Base & 36.78 & 26.83 \\
MoECE-GS-Base & 39.22 & 27.77 \\
MoECE-ST-Base & \textbf{39.37} & \textbf{27.90} \\

\hline
T5-v1.1-Large & 42.08 & 26.14 \\
MoECE-GS-Large & 42.82 & 27.14 \\
MoECE-ST-Large & \textbf{43.06} & \textbf{27.48} \\
\hline
\end{tabular}
\caption{\label{tab:gmeg}
MoECE achieves higher $F_{0.5}$ scores than the dense model on the CWEB-G and CWEB-S test sets with a relatively small overhead for the computation in the router.
}
\vspace*{-0.5\baselineskip}
\end{table}

\section{Results}
We report the scores of our models and the comparable dense model in Table~\ref{tab:bea_conll} and Table~\ref{tab:gmeg}. Our MoECE-GS-Base model successfully improves the $F_{0.5}$ score by 0.43 points on the CoNLL-2014 test set and 1.32 points on the BEA-2019 test set, while our MoECE-ST-Base improves the $F_{0.5}$ score by 0.55 points on the CoNLL-2014 test set and 0.85 points on the BEA-2019 test set. The improvements are statistically significant with $p < 0.01$.

\begin{figure*}[thb]
\centering
\begin{subfigure}{.5\textwidth}
  \centering
  \includegraphics[width=.94\linewidth]{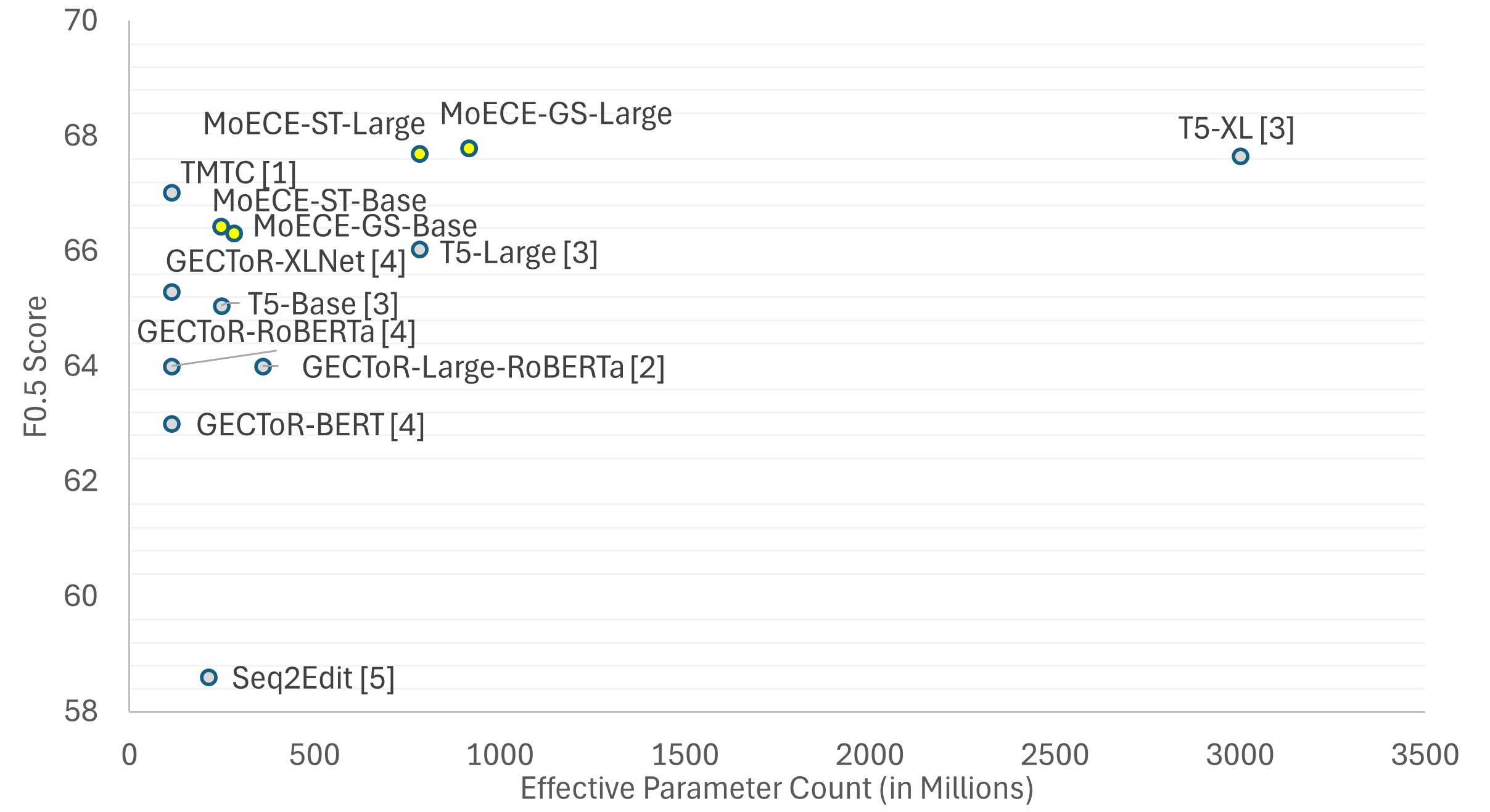}
  \caption{CoNLL-2014 test set}
  \label{fig:sub1}
\end{subfigure}%
\begin{subfigure}{.5\textwidth}
  \centering
  \includegraphics[width=.94\linewidth]{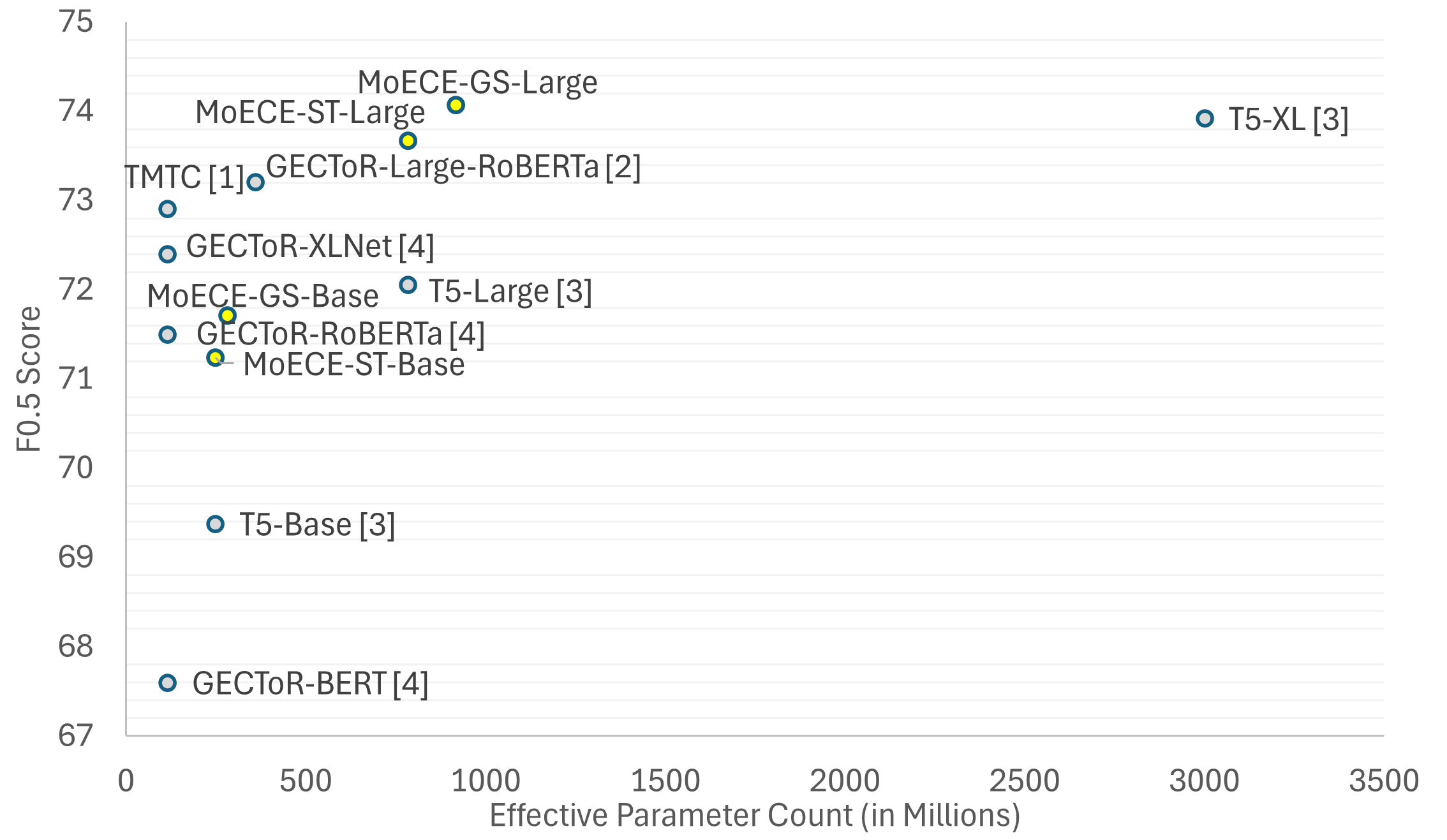}
  \caption{BEA-2019 test set}
  \label{fig:sub2}
\end{subfigure}
\caption{$F_{0.5}$ scores of comparable models that produce the error type of corrections according to the effective parameter count (in millions). Legends: [1] \citet{lai-etal-2022-type} [2] \citet{tarnavskyi-etal-2022-ensembling_short}, [3] \citet{rothe-etal-2021-simple_short}, [4] \citet{omelianchuk-etal-2020-gector_short}, [5] \citet{stahlberg-kumar-2020-seq2edits_short}. We do not compare against \cite{sun-wang-2022-adjusting} and \cite{bout-etal-2023-efficient} which only produce the correction tokens.
}
\label{fig:result_prev_work}
\end{figure*}

Without re-training or changing the hyper-parameters, MoECE also brings improvements over the dense model on out-of-domain datasets (Table \ref{tab:gmeg}). MoECE-GS-Base improves the $F_{0.5}$ score on the CWEB-G test set and CWEB-S test set by 2.44 points and 0.94 points respectively, while MoECE-ST-Base improves the $F_{0.5}$ score on the same test sets by 2.59 points and 1.07 points. Our method still improves the $F_{0.5}$ scores on all test sets when we scale up the models to MoECE-GS-Large and MoECE-ST-Large, and the improvements are statistically significant with $p < 0.01$.

Compared to previous work (Figure~\ref{fig:result_prev_work}), the performance of our MoECE-GS-Large model is comparable to T5-XL \cite{rothe-etal-2021-simple_short} despite having vastly fewer effective parameters. T5-XL with ~3B effective parameters has an $F_{0.5}$ score of 67.65\footnote{We use the score from the updated paper on \href{https://arxiv.org/abs/2106.03830}{arXiv}.} on the CoNLL-2014 test set and 73.92 on the BEA-2019 test set, while MoECE-GS-Large that has three times fewer effective parameters has an $F_{0.5}$ score of 67.79 on the CoNLL-2014 test set and 74.07 on the BEA-2019 test set (Table~\ref{tab:bea_conll}). Even if we include the unused experts, MoECE-GS-Large with 1.7B parameters is still smaller than T5-XL. This shows that our model is more efficient in terms of computation and memory usage than T5-XL.

\section{Analysis}

\begin{table*}
\centering
\begin{tabular}
{l c c | r r r r | r}
\hline
\textbf{Router} & \textbf{LB} & \textbf{ET} & \textbf{CoNLL-2014} & \textbf{BEA-2019} & \textbf{CWEB-G} & \textbf{CWEB-S} & \textbf{Avg}\\
\hline
GShard & \cmark & \cmark & 66.31 & 71.71 & \textbf{39.22} & 27.77 & \textbf{51.25}\\
GShard & \cmark & \xmark & \textbf{66.34} & \textbf{71.80} & 38.84 & \textbf{27.88} & 51.22 \\
\hline
SwitchTransformer & \cmark & \cmark & \textbf{66.43} & \textbf{71.24} & \textbf{39.37} & 27.90 & \textbf{51.24}\\
SwitchTransformer & \cmark & \xmark & 66.31 & 69.72 & 37.97 & \textbf{27.97} & 50.49 \\
\hline
GShard & \xmark & \cmark & \textbf{66.23} & \textbf{71.54} & \textbf{39.39} & \textbf{28.55} & \textbf{51.43} \\

\hline
\end{tabular}
\caption{\label{tab:edit_loss}
The effect of the load balancing loss (LB) and error type loss (ET) when training MoECE-Base. The evaluation is based on the $F_{0.5}$ scores on the CoNLL-2014 test set, BEA-2019 test set, CWEB-G test set, CWEB-S test set, and the average of these four test sets.
}
\end{table*}

\subsection{Routing Policy}
\label{sec:routing_policy}
\begin{figure}[htb]
\centering
\includegraphics[width=0.99\linewidth]
{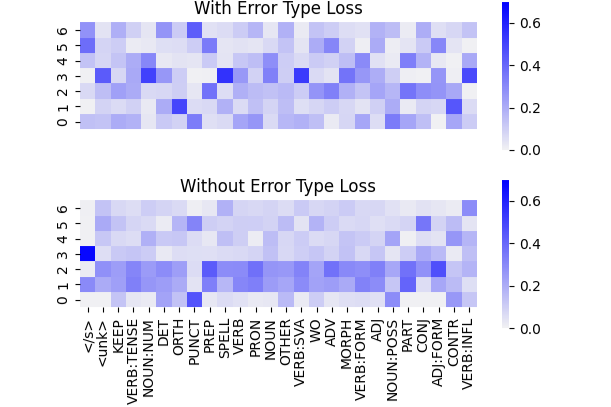}
\caption{Average routing score of MoECE-GS-Base for each token based on the error type.}
\label{fig:gate-distribution}
\end{figure}

We analyze whether the error type loss helps the router in choosing the experts based on the error type of the token. We take the average routing score that the last router produces for each error type. We find that with the error type loss, the router chooses different combinations of experts according to the token error type. In Figure~\ref{fig:gate-distribution}, we can see that the router mainly chooses experts \#0 and \#6 when correcting punctuation errors (\texttt{PUNCT}) and experts \#2 and \#5 when correcting preposition errors (\texttt{PREP}). On the other hand, when the error type loss is not used, experts \#1 and \#2 dominate the routing policy for correcting almost any error type.

\begin{figure}[htb]
\centering
\includegraphics[width=0.9\linewidth]
{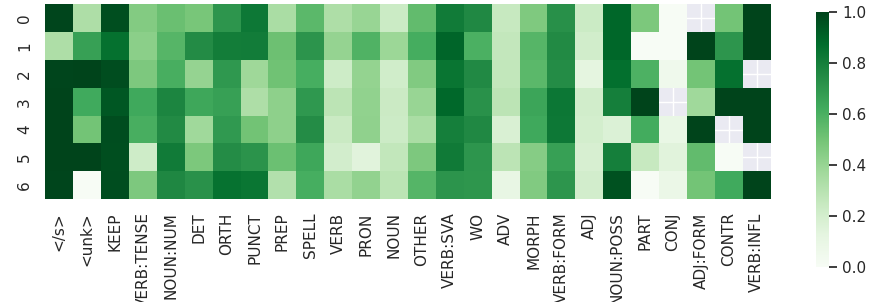}
\caption{Experts' correction accuracy for each token based on the error type.}
\label{fig:expert-acc}
\end{figure}

The routing process also leads to the specialization of experts (Figure~\ref{fig:expert-acc}). For example, in correcting punctuation errors, we observe that expert \#6 and expert \#0 achieve the highest accuracy, at 83.7\% and 82.6\%, respectively. In contrast, the accuracy of expert \#3, expert \#2, and expert \#4 is 33.3\%, 38.1\%, and 50.7\%, respectively. Similarly, for orthographical errors, expert \#1 achieves 80.9\% accuracy, while expert \#3 achieves 66.7\% accuracy. The accuracy of each expert aligns well with the routing policy, especially for error types frequently encountered in the training data, with the median Pearson correlation coefficient being 0.66 for the top seven error types.

We evaluate the accuracy of error type prediction of the router on the BEA-2019 development set and find that the overall accuracy of error type prediction by the routers is 92.5\%. This provides evidence that the model is capable of learning the error type using the classification head in the router and suggests that the model utilizes this information to route the tokens to the appropriate experts.

\subsection{Impact of Error Type Loss and Load Balancing Loss}
Despite having a more comprehensible routing decision, models trained with the error type loss do not strictly perform better than models trained without it (Table~\ref{tab:edit_loss}). This shows that the routing can learn a different routing policy by itself that performs comparably well, even though the policy is not based on the error type. However, the routing policy may be hard to interpret and subsequently makes it harder to understand if the model makes weird mistakes. On the other hand, the model trained with the error type loss will produce both the corrected text and the error type at the same time during inference. This will help language learners understand the reason for the correction and can help model developers look for the cause of the issue if the model makes mistakes. On top of that, the expert allocation for each error type is quite clear when the model is trained with error type loss (Figure~\ref{fig:gate-distribution}). This allows some degree of modularity to add, modify, or remove the experts during deployment.

We also run an experiment of training the model with the error type loss but without the load balancing loss. In this experimental setting, we use the same routing criteria as GShard but set the load balancing loss multiplier $\beta$ (Equation~\ref{eq:final_loss}) to zero. We also obtain a comparable model to the other experimental settings. This shows that the error type loss can serve as an alternative to the existing load balancing loss.

\begin{table*}
\centering
\begin{tabular}{l | l | r | c | c | r r r | r r r }
\hline
{} & {} & \textbf{Exp} & {} & \textbf{BEA-2019} & \multicolumn{3}{c|}{\textbf{CoNLL-2014 Test}} & \multicolumn{3}{c}{\textbf{BEA-2019 Test}}\\
{\#} & \textbf{Model} & \textbf{Dim} & \textbf{EPC} & \textbf{Dev (F\textsubscript{0.5})} & \textbf{P} & \textbf{R} & \textbf{F\textsubscript{0.5}} & \textbf{P} & \textbf{R} & \textbf{F\textsubscript{0.5}}\\
\hline
1 & T5-v1.1-Base & - & 248M & 53.97 & 72.43 & 48.38 & 65.88 & 72.82 & 62.11 & 70.39\\
\hline
2 & MoECE-ST & 2048 & 248M & 54.84 & 72.75 & \textbf{49.31} & \textbf{66.43} & 73.99 & 62.00 & 71.24 \\
3 & MoECE-ST & 128 & 250M & 54.86 & 72.96 & 48.51 & 66.28 & 74.56 & 61.37 & 71.49 \\
4 & MoECE-GS & 128 & 252M & 54.86 & 72.18 & 49.22 & 66.02 & 73.74 & \textbf{62.42} & 71.16 \\
5 & MoECE-GS & 2048 & 282M & \textbf{55.28} & \textbf{73.00} & 48.53 & 66.31 & \textbf{74.59} & 62.11 & \textbf{71.71} \\
\hline
\end{tabular}
\caption{\label{tab:low_rank}
Scores of MoECE-GS(-Base) and MoECE-ST(-Base) with original and low-rank expert dimensions (Exp Dim) on the BEA-2019 development set (Dev), CoNLL-2014 test set, and BEA-2019 test set. The rows are sorted from the models with the lowest effective parameter count (EPC) to the highest. The low-rank MoECE-ST-Base has a higher EPC than the original because the original merges the shared feed-forward layer with the feed-forward layers in each expert after training.
}
\end{table*}

\subsection{Shared Feed-Forward Layer}
\label{sec:shared_layer}
As explained in Section \ref{sec:moe_layer}, we design our MoE to be an addition to the main transformer feed-forward layer, rather than a replacement, during training. We hypothesize that for a neural network to correct errors in text, it requires general skills and specific skills according to the error types. By adding the output of the MoE with the main feed-forward layer, the general skills can be learned by the feed-forward layer that is shared across all experts. This idea is similar to DeepSeekMoE \cite{dai2024deepseekmoe}, in which some experts are always chosen for any input token so that those experts learn the necessary general skills.

\begin{figure}[bht]
\centering
\includegraphics[width=0.99\linewidth]
{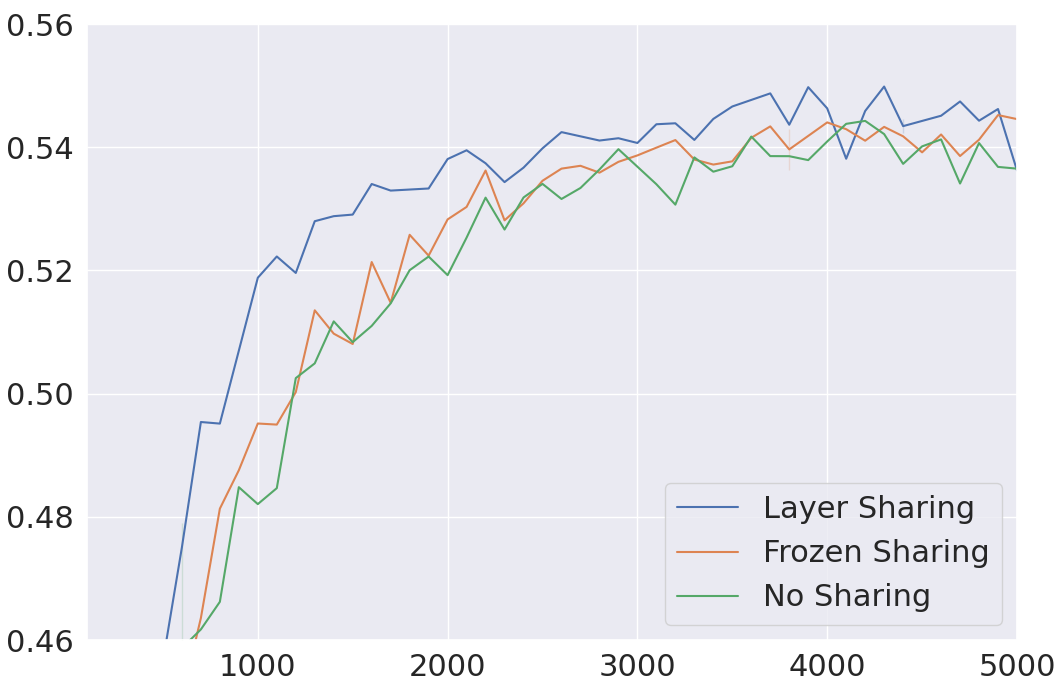}
\caption{$F_{0.5}$ scores on the BEA-2019 development set of three architecture types: MoECE (\textsc{Layer Sharing}), MoECE without parameter updates to the shared layer (\textsc{Frozen Sharing}), and standard MoE architecture where the feed-forward layer is entirely replaced by the MoE layer (\textsc{No Sharing}).}
\label{fig:shared}
\end{figure}

We investigate the efficacy of our architecture design by comparing three architecture types: our MoECE architecture (\textsc{Layer Sharing}), our MoECE architecture without parameter updates to the shared layer (\textsc{Frozen Sharing}), and the common mixture-of-experts architecture where the feed-forward layer is replaced by the MoE layer (\textsc{No Sharing}). With no gradient updates to the shared layer, the shared skill between experts is limited to what was previously learned during the pre-training of the language model. On the other hand, when there is no layer sharing, each expert must learn the general skills independently without any information sharing.

In the layer sharing and frozen sharing experimental settings, the feed-forward layer is initialized with the pre-trained weights from the language model while the experts are initialized randomly. In the experiment without layer sharing, the feed-forward layer is removed (replaced entirely by the MoE layer), so the weights of the feed-forward layer are used to initialize the experts in the MoE layer. Small random noise is added to the expert weights to encourage the experts to specialize in different aspects rather than becoming copies of the same network.

We observe that the model with frozen layer sharing produces a similar $F_{0.5}$ score on the BEA-2019 development set after convergence, but it takes much longer to train. As seen in Figure~\ref{fig:shared}, the model with a shared layer reaches an $F_{0.5}$ score of 54\% after 2,100 parameter updates, while the model with a frozen feed-forward layer requires 3,100 parameter updates. On the other hand, removing the shared layer entirely results in a lower score after convergence. 
Having a shared feed-forward layer in the transformer block adds insignificant computational costs during training, and the costs disappear after merging the layer with each expert during inference.

\subsection{Low-Rank Experts}
The MoECE architecture bears some resemblance to the Low-Rank Adaptation (LoRA) network \cite{hu2022lora}. In MoECE’s architecture, the dimensions of the feed-forward layers in the experts are the same as those of the main transformer feed-forward layer. We investigate whether similar performance can be achieved by using experts with feed-forward layers that project the hidden representation into much smaller dimensions and then project back to the original dimensions, similar to LoRA. Using this type of expert network reduces memory usage and disk space, but the main transformer feed-forward layer can no longer be merged into the expert layers due to the difference in dimensions.

In Table~\ref{tab:low_rank}, we observe that while MoECE-GS with low-rank experts (row \#4) does not perform as well, MoECE-ST with low-rank experts (row \#3) performs comparably to the original MoECE-ST (row \#2), with slightly higher $F_{0.5}$ score on the BEA-2019 test set but lower $F_{0.5}$ score on the CoNLL-2014 test set. This indicates that low-rank experts are a good alternative when memory consumption is a concern.

\section{Related Work}
In this section, we briefly outline previous work on interpretable grammatical error correction and recent work on mixture of experts for transformer models.
\subsection{Interpretable GEC}
An interpretable GEC model is desirable as it can help the user understand the rationale behind the correction produced by a GEC model. Since many GEC users are language learners, understanding the rationale behind the corrections can help them avoid making the same mistakes in the future. \citet{stahlberg-kumar-2020-seq2edits_short} propose a model that generates error types during inference, but the main output of the model is a sequence of edits instead of the corrected text. Sequence-tagging models such as GECToR \cite{omelianchuk-etal-2020-gector_short} also provide interpretability from the general transformation tags that it outputs, but those are only a small subset of the output tags. Showing the error types may not be considered complete feedback, but error types can still serve as useful feedback to the user \cite{qorib-etal-2023-allecs}.

\citet{kaneko-etal-2022-interpretability} propose a GEC system that produces a corrected text with an example of a similar sentence from the training data to make their model interpretable. \citet{fei-etal-2023-enhancing} propose a GEC system that accompanies the corrections with the error types and the evidence words. Even so, these approaches still have their own weaknesses. More research is needed on interpretable and explainable grammatical error correction to provide more useful feedback to language learners.

\subsection{Mixture of Experts}
Mixture of experts is a classic method that was proposed by \citet{Jacobs1991AdaptiveMO} and reinterpreted for neural networks by \citet{shazeer2017}. \citet{shazeer2017} reach the state of the art on language modeling and machine translation by applying MoE convolutionally between stacked LSTM layers.

MoE has recently gained popularity thanks to its effectiveness in scaling up the number of parameters of transformer models while maintaining reasonable computation cost \cite{Fedus2022Review}. Much research has been done on improving training stability \cite{Du2021GLaMES, zoph2022stmoe}, the router \cite{NEURIPS2022_2f00ecd7, Lewis2021BASELS}, and the load balancing loss \cite{lepikhin2021gshard, 10.5555/3586589.3586709}, but little has been done on its transferability from dense models. Most MoE models go through a pre-training process instead of transferring the weights from a dense model and transforming it into an MoE during fine-tuning, which is what we have done in this work. \citet{gao-etal-2022-parameter} propose to expand a pre-trained language model into a mixture of experts, but they use the parameter matrix of the matrix product operator (a tensor decomposition from quantum many-body physics) to be the expert. Their architecture design is significantly different from ours which uses the MoE layer as an addition instead of a replacement of the transformer feed-forward layer.

\section{Conclusion and Future Work}

In this paper, we present a new grammatical error correction model that is more efficient by utilizing a mixture of experts, called MoECE. Our experiments show that our model can improve the $F_{0.5}$ scores of the comparable dense model by up to 0.55 points on the CoNLL-2014 test set and 1.32 points on the BEA-2019 test set. With the same model and hyper-parameters, the model can improve the $F_{0.5}$ score on out-of-domain test sets by up to 2.59 points on CWEB-G and 1.07 points on CWEB-S. The larger variant of our model, MoECE-GS-Large, successfully reaches performance slightly better than a model, T5-XL, that has three times its effective parameter count and almost double its total parameter count.

Our proposed error type loss makes our model interpretable by producing corrections with error types. Our analysis shows that the error type loss helps in routing the input token to the appropriate expert based on its error type. In addition, we find that our error type loss can be an alternative to existing load balancing loss. We believe that interpretable and explainable grammatical error correction models are needed to help language learners with their study and we hope more research explores this direction.

\section*{Limitations}
In this work, we only investigate grammatical error correction for English. Our method is applicable to grammatical error correction for other languages when sufficient training data is available. We have not run experiments on larger models due to limitation of our compute budget, but we believe our current experimental configurations are sufficient to empirically demonstrate the effectiveness of our method. We believe our work does not bring any direct harm to individuals or society.
\bibliography{custom}

\begin{thebibliography}{43}
\providecommand{\natexlab}[1]{#1}

\bibitem[{Bout et~al.(2023)Bout, Podolskiy, Nikolenko, and Piontkovskaya}]{bout-etal-2023-efficient}
Andrey Bout, Alexander Podolskiy, Sergey Nikolenko, and Irina Piontkovskaya. 2023.
\newblock \href {https://doi.org/10.18653/v1/2023.emnlp-main.355} {Efficient grammatical error correction via multi-task training and optimized training schedule}.
\newblock In \emph{Proceedings of EMNLP}, pages 5800--5816.

\bibitem[{Bryant et~al.(2019)Bryant, Felice, Andersen, and Briscoe}]{bryant-etal-2019-bea_short}
Christopher Bryant, Mariano Felice, {\O}istein~E. Andersen, and Ted Briscoe. 2019.
\newblock \href {https://doi.org/10.18653/v1/W19-4406} {The {BEA}-2019 shared task on grammatical error correction}.
\newblock In \emph{Proceedings of BEA}, pages 52--75.

\bibitem[{Bryant et~al.(2017)Bryant, Felice, and Briscoe}]{errant}
Christopher Bryant, Mariano Felice, and Ted Briscoe. 2017.
\newblock \href {https://doi.org/10.18653/v1/P17-1074} {Automatic annotation and evaluation of error types for grammatical error correction}.
\newblock In \emph{Proceedings of ACL}, pages 793--805.

\bibitem[{Bryant et~al.(2023)Bryant, Yuan, Qorib, Cao, Ng, and Briscoe}]{gec_survey}
Christopher Bryant, Zheng Yuan, Muhammad~Reza Qorib, Hannan Cao, Hwee~Tou Ng, and Ted Briscoe. 2023.
\newblock \href {https://doi.org/10.1162/coli_a_00478} {Grammatical error correction: A survey of the state of the art}.
\newblock \emph{Computational Linguistics}, 49(3):643--701.

\bibitem[{Chollampatt and Ng(2018)}]{10.5555/3504035.3504741}
Shamil Chollampatt and Hwee~Tou Ng. 2018.
\newblock \href {https://dl.acm.org/doi/10.5555/3504035.3504741} {A multilayer convolutional encoder-decoder neural network for grammatical error correction}.
\newblock In \emph{Proceedings of AAAI}, pages 5755--5762.

\bibitem[{Chollampatt et~al.(2016)Chollampatt, Taghipour, and Ng}]{chollampatt-etal-2016-neural}
Shamil Chollampatt, Kaveh Taghipour, and Hwee~Tou Ng. 2016.
\newblock \href {https://www.ijcai.org/Proceedings/16/Papers/393.pdf} {Neural network translation models for grammatical error correction}.
\newblock In \emph{Proceedings of IJCAI}, pages 2768--2774.

\bibitem[{Dahlmeier and Ng(2012)}]{dahlmeier-ng-2012-better_short}
Daniel Dahlmeier and Hwee~Tou Ng. 2012.
\newblock \href {https://aclanthology.org/N12-1067} {Better evaluation for grammatical error correction}.
\newblock In \emph{Proceedings of NAACL}, pages 568--572.

\bibitem[{Dai et~al.(2024)Dai, Deng, Zhao, Xu, Gao, Chen, Li, Zeng, Yu, Wu, Xie, Li, Huang, Luo, Ruan, Sui, and Liang}]{dai2024deepseekmoe}
Damai Dai, Chengqi Deng, Chenggang Zhao, R.~X. Xu, Huazuo Gao, Deli Chen, Jiashi Li, Wangding Zeng, Xingkai Yu, Y.~Wu, Zhenda Xie, Y.~K. Li, Panpan Huang, Fuli Luo, Chong Ruan, Zhifang Sui, and Wenfeng Liang. 2024.
\newblock \href {https://arxiv.org/abs/2401.06066} {{DeepSeekMoE}: Towards ultimate expert specialization in mixture-of-experts language models}.
\newblock \emph{ArXiv}, abs/2401.06066.

\bibitem[{Du et~al.(2021)Du, Huang, Dai, Tong, Lepikhin, Xu, Krikun, Zhou, Yu, Firat, Zoph, Fedus, Bosma, Zhou, Wang, Wang, Webster, Pellat, Robinson, Meier-Hellstern, Duke, Dixon, Zhang, Le, Wu, Chen, and Cui}]{Du2021GLaMES}
Nan Du, Yanping Huang, Andrew~M. Dai, Simon Tong, Dmitry Lepikhin, Yuanzhong Xu, Maxim Krikun, Yanqi Zhou, Adams~Wei Yu, Orhan Firat, Barret Zoph, Liam Fedus, Maarten Bosma, Zongwei Zhou, Tao Wang, Yu~Emma Wang, Kellie Webster, Marie Pellat, Kevin Robinson, Kathleen~S. Meier-Hellstern, Toju Duke, Lucas Dixon, Kun Zhang, Quoc~V. Le, Yonghui Wu, Z.~Chen, and Claire Cui. 2021.
\newblock \href {https://api.semanticscholar.org/CorpusID:245124124} {Glam: Efficient scaling of language models with mixture-of-experts}.
\newblock In \emph{Proceedings of ICML}, pages 5547--5569.

\bibitem[{Fedus et~al.(2022{\natexlab{a}})Fedus, Dean, and Zoph}]{Fedus2022Review}
William Fedus, Jeff Dean, and Barret Zoph. 2022{\natexlab{a}}.
\newblock \href {https://arxiv.org/abs/2209.01667} {A review of sparse expert models in deep learning}.
\newblock \emph{ArXiv}, abs/2209.01667.

\bibitem[{Fedus et~al.(2022{\natexlab{b}})Fedus, Zoph, and Shazeer}]{10.5555/3586589.3586709}
William Fedus, Barret Zoph, and Noam Shazeer. 2022{\natexlab{b}}.
\newblock \href {https://jmlr.org/papers/v23/21-0998.html} {{Switch Transformers}: scaling to trillion parameter models with simple and efficient sparsity}.
\newblock \emph{JMLR}, 23(1):1--39.

\bibitem[{Fei et~al.(2023)Fei, Cui, Yang, Lam, Lan, and Shi}]{fei-etal-2023-enhancing}
Yuejiao Fei, Leyang Cui, Sen Yang, Wai Lam, Zhenzhong Lan, and Shuming Shi. 2023.
\newblock \href {https://doi.org/10.18653/v1/2023.acl-long.413} {Enhancing grammatical error correction systems with explanations}.
\newblock In \emph{Proceedings of ACL}, pages 7489--7501.

\bibitem[{Flachs et~al.(2020)Flachs, Lacroix, Yannakoudakis, Rei, and S{\o}gaard}]{flachs-etal-2020-grammatical_short}
Simon Flachs, Oph{\'e}lie Lacroix, Helen Yannakoudakis, Marek Rei, and Anders S{\o}gaard. 2020.
\newblock \href {https://doi.org/10.18653/v1/2020.emnlp-main.680} {Grammatical error correction in low error density domains: A new benchmark and analyses}.
\newblock In \emph{Proceedings of EMNLP}, pages 8467--8478.

\bibitem[{Gao et~al.(2022)Gao, Liu, Zhao, Lu, and Wen}]{gao-etal-2022-parameter}
Ze-Feng Gao, Peiyu Liu, Wayne~Xin Zhao, Zhong-Yi Lu, and Ji-Rong Wen. 2022.
\newblock \href {https://aclanthology.org/2022.coling-1.288} {Parameter-efficient mixture-of-experts architecture for pre-trained language models}.
\newblock In \emph{Proceedings of COLING}, pages 3263--3273.

\bibitem[{Han and Ng(2021)}]{ddc}
Wenjuan Han and Hwee~Tou Ng. 2021.
\newblock \href {https://arxiv.org/abs/2110.15149} {Diversity-driven combination for grammatical error correction}.
\newblock In \emph{Proceedings of ICTAI}, pages 972--979.

\bibitem[{He et~al.(2021)He, Qiu, Zeng, Yang, Zhai, and Tang}]{he2021fastmoe}
Jiaao He, Jiezhong Qiu, Aohan Zeng, Zhilin Yang, Jidong Zhai, and Jie Tang. 2021.
\newblock \href {https://arxiv.org/} {{FastMoE}: A fast mixture-of-expert training system}.
\newblock \emph{ArXiv}, abs/2103.13262.

\bibitem[{Hu et~al.(2022)Hu, yelong shen, Wallis, Allen-Zhu, Li, Wang, Wang, and Chen}]{hu2022lora}
Edward~J Hu, yelong shen, Phillip Wallis, Zeyuan Allen-Zhu, Yuanzhi Li, Shean Wang, Lu~Wang, and Weizhu Chen. 2022.
\newblock \href {https://openreview.net/forum?id=nZeVKeeFYf9} {Lo{RA}: Low-rank adaptation of large language models}.
\newblock In \emph{Proceedings of ICLR}.

\bibitem[{Jacobs et~al.(1991)Jacobs, Jordan, Nowlan, and Hinton}]{Jacobs1991AdaptiveMO}
Robert~A. Jacobs, Michael~I. Jordan, Steven~J. Nowlan, and Geoffrey~E. Hinton. 1991.
\newblock \href {https://ieeexplore.ieee.org/document/6797059} {Adaptive mixtures of local experts}.
\newblock \emph{Neural Computation}, 3:79--87.

\bibitem[{Kaneko et~al.(2022)Kaneko, Takase, Niwa, and Okazaki}]{kaneko-etal-2022-interpretability}
Masahiro Kaneko, Sho Takase, Ayana Niwa, and Naoaki Okazaki. 2022.
\newblock \href {https://doi.org/10.18653/v1/2022.acl-long.496} {Interpretability for language learners using example-based grammatical error correction}.
\newblock In \emph{Proceedings of ACL}, pages 7176--7187.

\bibitem[{Kantor et~al.(2019)Kantor, Katz, Choshen, Cohen-Karlik, Liberman, Toledo, Menczel, and Slonim}]{kantor-etal-2019-learning}
Yoav Kantor, Yoav Katz, Leshem Choshen, Edo Cohen-Karlik, Naftali Liberman, Assaf Toledo, Amir Menczel, and Noam Slonim. 2019.
\newblock \href {https://doi.org/10.18653/v1/W19-4414} {Learning to combine grammatical error corrections}.
\newblock In \emph{Proceedings of BEA}, pages 139--148.

\bibitem[{Kudugunta et~al.(2021)Kudugunta, Huang, Bapna, Krikun, Lepikhin, Luong, and Firat}]{kudugunta-etal-2021-beyond-distillation_short}
Sneha Kudugunta, Yanping Huang, Ankur Bapna, Maxim Krikun, Dmitry Lepikhin, Minh-Thang Luong, and Orhan Firat. 2021.
\newblock \href {https://doi.org/10.18653/v1/2021.findings-emnlp.304} {Beyond distillation: Task-level mixture-of-experts for efficient inference}.
\newblock In \emph{Findings of EMNLP}, pages 3577--3599.

\bibitem[{Lai et~al.(2022)Lai, Zhou, Zeng, Li, Li, Cao, and Su}]{lai-etal-2022-type}
Shaopeng Lai, Qingyu Zhou, Jiali Zeng, Zhongli Li, Chao Li, Yunbo Cao, and Jinsong Su. 2022.
\newblock \href {https://doi.org/10.18653/v1/2022.findings-acl.254} {Type-driven multi-turn corrections for grammatical error correction}.
\newblock In \emph{Findings of ACL 2022}, pages 3225--3236.

\bibitem[{Lepikhin et~al.(2021)Lepikhin, Lee, Xu, Chen, Firat, Huang, Krikun, Shazeer, and Chen}]{lepikhin2021gshard}
Dmitry Lepikhin, HyoukJoong Lee, Yuanzhong Xu, Dehao Chen, Orhan Firat, Yanping Huang, Maxim Krikun, Noam Shazeer, and Zhifeng Chen. 2021.
\newblock \href {https://openreview.net/forum?id=qrwe7XHTmYb} {{GS}hard: Scaling giant models with conditional computation and automatic sharding}.
\newblock In \emph{Proceedings of ICLR}.

\bibitem[{Lewis et~al.(2021)Lewis, Bhosale, Dettmers, Goyal, and Zettlemoyer}]{Lewis2021BASELS}
Mike Lewis, Shruti Bhosale, Tim Dettmers, Naman Goyal, and Luke Zettlemoyer. 2021.
\newblock \href {https://api.semanticscholar.org/CorpusID:232428341} {Base layers: Simplifying training of large, sparse models}.
\newblock In \emph{Proceedings of ICML}, pages 6265--6274.

\bibitem[{Li et~al.(2020)Li, Li, Wang, and Li}]{li-etal-video_short}
Dingcheng Li, Xu~Li, Jun Wang, and Ping Li. 2020.
\newblock \href {https://doi.org/10.1145/3397271.3401238} {Video recommendation with multi-gate mixture of experts soft actor critic}.
\newblock In \emph{Proceedings of SIGIR}, pages 1553--1556.

\bibitem[{Li et~al.(2023)Li, Liu, Wang, Gong, Wong, Gao, Huang, and Zhang}]{li-etal-2023-templategec_short}
Yinghao Li, Xuebo Liu, Shuo Wang, Peiyuan Gong, Derek~F. Wong, Yang Gao, Heyan Huang, and Min Zhang. 2023.
\newblock \href {https://doi.org/10.18653/v1/2023.acl-long.380} {{T}emplate{GEC}: Improving grammatical error correction with detection template}.
\newblock In \emph{Proceedings of ACL}, pages 6878--6892.

\bibitem[{Lin and Ng(2021)}]{gec-ip}
Ruixi Lin and Hwee~Tou Ng. 2021.
\newblock \href {https://aclanthology.org/2021.ranlp-1.94} {System combination for grammatical error correction based on integer programming}.
\newblock In \emph{Proceedings of RANLP}, pages 829--834.

\bibitem[{Omelianchuk et~al.(2020)Omelianchuk, Atrasevych, Chernodub, and Skurzhanskyi}]{omelianchuk-etal-2020-gector_short}
Kostiantyn Omelianchuk, Vitaliy Atrasevych, Artem Chernodub, and Oleksandr Skurzhanskyi. 2020.
\newblock \href {https://doi.org/10.18653/v1/2020.bea-1.16} {{GECT}o{R} {--} grammatical error correction: Tag, not rewrite}.
\newblock In \emph{Proceedings of BEA}, pages 163--170.

\bibitem[{Ott et~al.(2019)Ott, Edunov, Baevski, Fan, Gross, Ng, Grangier, and Auli}]{ott2019fairseq}
Myle Ott, Sergey Edunov, Alexei Baevski, Angela Fan, Sam Gross, Nathan Ng, David Grangier, and Michael Auli. 2019.
\newblock \href {https://doi.org/10.18653/v1/N19-4009} {fairseq: A fast, extensible toolkit for sequence modeling}.
\newblock In \emph{Proceedings of NAACL-HLT 2019: Demonstrations}, pages 48--53.

\bibitem[{Qorib et~al.(2023)Qorib, Moon, and Ng}]{qorib-etal-2023-allecs}
Muhammad~Reza Qorib, Geonsik Moon, and Hwee~Tou Ng. 2023.
\newblock \href {https://doi.org/10.18653/v1/2023.eacl-demo.32} {{ALLECS}: A lightweight language error correction system}.
\newblock In \emph{Proceedings of EACL}, pages 298--306.

\bibitem[{Qorib et~al.(2022)Qorib, Na, and Ng}]{qorib-etal-2022-frustratingly_short}
Muhammad~Reza Qorib, Seung-Hoon Na, and Hwee~Tou Ng. 2022.
\newblock \href {https://doi.org/10.18653/v1/2022.naacl-main.143} {Frustratingly easy system combination for grammatical error correction}.
\newblock In \emph{Proceedings of NAACL}, pages 1964--1974.

\bibitem[{Qorib and Ng(2023)}]{qorib-ng-2023-system_short}
Muhammad~Reza Qorib and Hwee~Tou Ng. 2023.
\newblock \href {https://doi.org/10.18653/v1/2023.emnlp-main.785} {System combination via quality estimation for grammatical error correction}.
\newblock In \emph{Proceedings of EMNLP}, pages 12746--12759.

\bibitem[{Raffel et~al.(2020)Raffel, Shazeer, Roberts, Lee, Narang, Matena, Zhou, Li, and Liu}]{JMLR:t5}
Colin Raffel, Noam Shazeer, Adam Roberts, Katherine Lee, Sharan Narang, Michael Matena, Yanqi Zhou, Wei Li, and Peter~J. Liu. 2020.
\newblock \href {http://jmlr.org/papers/v21/20-074.html} {Exploring the limits of transfer learning with a unified text-to-text transformer}.
\newblock \emph{JMLR}, 21(140):1--67.

\bibitem[{Rothe et~al.(2021)Rothe, Mallinson, Malmi, Krause, and Severyn}]{rothe-etal-2021-simple_short}
Sascha Rothe, Jonathan Mallinson, Eric Malmi, Sebastian Krause, and Aliaksei Severyn. 2021.
\newblock \href {https://doi.org/10.18653/v1/2021.acl-short.89} {A simple recipe for multilingual grammatical error correction}.
\newblock In \emph{Proceedings of ACL}, pages 702--707.

\bibitem[{Shazeer et~al.(2017)Shazeer, Mirhoseini, Maziarz, Davis, Le, Hinton, and Dean}]{shazeer2017}
Noam Shazeer, Azalia Mirhoseini, Krzysztof Maziarz, Andy Davis, Quoc Le, Geoffrey Hinton, and Jeff Dean. 2017.
\newblock \href {https://openreview.net/forum?id=B1ckMDqlg} {Outrageously large neural networks: The sparsely-gated mixture-of-experts layer}.
\newblock In \emph{Proceedings of ICLR}.

\bibitem[{Sorokin(2022)}]{sorokin-2022-improved_short}
Alexey Sorokin. 2022.
\newblock \href {https://doi.org/10.18653/v1/2022.emnlp-main.785} {Improved grammatical error correction by ranking elementary edits}.
\newblock In \emph{Proceedings of EMNLP}, pages 11416--11429.

\bibitem[{Stahlberg and Kumar(2020)}]{stahlberg-kumar-2020-seq2edits_short}
Felix Stahlberg and Shankar Kumar. 2020.
\newblock \href {https://doi.org/10.18653/v1/2020.emnlp-main.418} {{S}eq2{E}dits: Sequence transduction using span-level edit operations}.
\newblock In \emph{Proceedings of EMNLP}, pages 5147--5159.

\bibitem[{Sun and Wang(2022)}]{sun-wang-2022-adjusting}
Xin Sun and Houfeng Wang. 2022.
\newblock \href {https://doi.org/10.18653/v1/2022.acl-short.77} {Adjusting the precision-recall trade-off with align-and-predict decoding for grammatical error correction}.
\newblock In \emph{Proceedings of ACL}, pages 686--693.

\bibitem[{Susanto et~al.(2014)Susanto, Phandi, and Ng}]{susanto-etal-2014-system_short}
Raymond~Hendy Susanto, Peter Phandi, and Hwee~Tou Ng. 2014.
\newblock \href {https://doi.org/10.3115/v1/D14-1102} {System combination for grammatical error correction}.
\newblock In \emph{Proceedings of {EMNLP}}, pages 951--962.

\bibitem[{Tarnavskyi et~al.(2022)Tarnavskyi, Chernodub, and Omelianchuk}]{tarnavskyi-etal-2022-ensembling_short}
Maksym Tarnavskyi, Artem Chernodub, and Kostiantyn Omelianchuk. 2022.
\newblock \href {https://doi.org/10.18653/v1/2022.acl-long.266} {Ensembling and knowledge distilling of large sequence taggers for grammatical error correction}.
\newblock In \emph{Proceedings of ACL}, pages 3842--3852.

\bibitem[{Yuan et~al.(2019)Yuan, Stahlberg, Rei, Byrne, and Yannakoudakis}]{yuan-etal-2019-neural_short}
Zheng Yuan, Felix Stahlberg, Marek Rei, Bill Byrne, and Helen Yannakoudakis. 2019.
\newblock \href {https://doi.org/10.18653/v1/W19-4424} {Neural and {FST}-based approaches to grammatical error correction}.
\newblock In \emph{Proceedings of BEA}, pages 228--239.

\bibitem[{Zhou et~al.(2022)Zhou, Lei, Liu, Du, Huang, Zhao, Dai, Chen, Le, and Laudon}]{NEURIPS2022_2f00ecd7}
Yanqi Zhou, Tao Lei, Hanxiao Liu, Nan Du, Yanping Huang, Vincent Zhao, Andrew~M Dai, zhifeng Chen, Quoc~V Le, and James Laudon. 2022.
\newblock \href {https://proceedings.neurips.cc/paper_files/paper/2022/file/2f00ecd787b432c1d36f3de9800728eb-Paper-Conference.pdf} {Mixture-of-experts with expert choice routing}.
\newblock In \emph{Proceedings of NeurIPS}, pages 7103--7114.

\bibitem[{Zoph et~al.(2022)Zoph, Bello, Kumar, Du, Huang, Dean, Shazeer, and Fedus}]{zoph2022stmoe}
Barret Zoph, Irwan Bello, Sameer Kumar, Nan Du, Yanping Huang, Jeff Dean, Noam Shazeer, and William Fedus. 2022.
\newblock \href {https://arxiv.org/abs/2202.08906} {{ST-MoE}: Designing stable and transferable sparse expert models}.
\newblock \emph{ArXiv}, abs/2202.08906.

\end{thebibliography}

\appendix

\begin{figure*}[tb]
\centering
\includegraphics[width=0.9\linewidth]{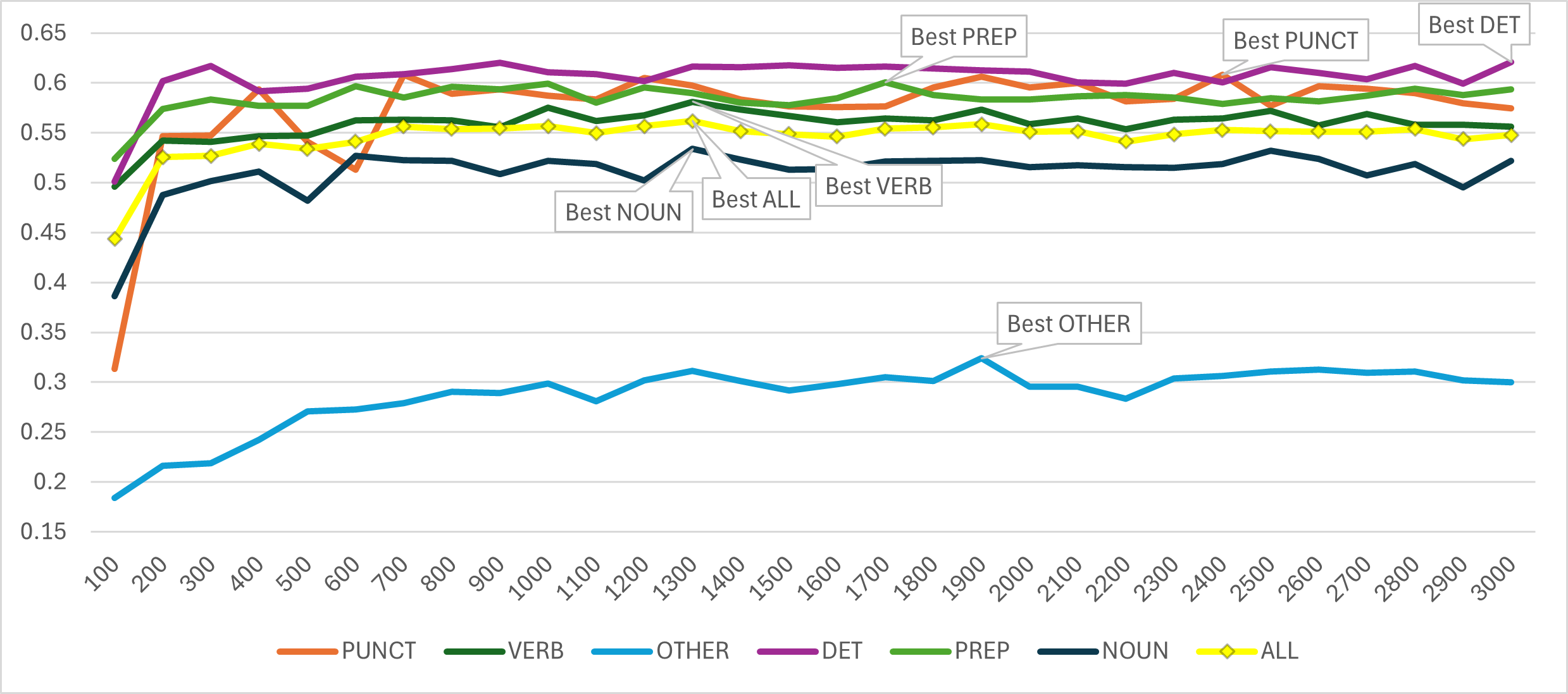}
\caption{$F_{0.5}$ scores of a T5-v1.1-Large model on different error types in the BEA-2019 development set at different numbers of training steps.}\label{fig:t5-large}
\end{figure*}

\section{Task Interference in T5-v1.1-Large}
\label{sec:ap_t5_large}
We observe a similar indication of task interference that we explain in the introduction on the training of T5-v1.1-Large (Figure~\ref{fig:t5-large}).

\section{Compute Budget}
\label{ap:compute}
We list the total number of parameters of our models in Table \ref{tab:parameter}. The training of the base models took about 33.3 hours for each model on a single NVIDIA H100 GPU, while training the large models took about 16.5 hours on two NVIDIA H100 GPUs. The reason why the training times of the base and large models are similar is that large models converged much earlier.

\begin{table}[thb]
\centering
\begin{tabular}{l p{0.06\linewidth} | p{0.1\linewidth} p{0.1\linewidth} }
\hline
Model & $M$ & EPC & TPC\\
\hline
MoECE-GS-Base & 7 & 282M & 490M \\
MoECE-ST-Base & 7 & 248M & 490M \\
MoECE-GS-Large & 7 & 917M & 1.7B \\
MoECE-ST-Large & 7 & 784M & 1.7B \\
\hline
\end{tabular}
\caption{\label{tab:parameter}
Effective parameter counts (EPC) and total parameter counts (TPC) of our models. 
}
\end{table}

\section{Experiments}
The parameters that we need to set to train the models are given in Table \ref{tab:hyper-param}. The rest follows the default hyper-parameters of Fairseq\footnote{\url{https://github.com/facebookresearch/fairseq/}} \cite{ott2019fairseq}. Testing the model does not require specifying the hyper-parameters. We did not conduct an extensive hyper-parameter search, but we specify our parameter bounds in Table \ref{tab:hp-bound}. Note that we perform hyper-parameter search on the smaller models.

We performed experiments on multiple configurations to verify the effectiveness of our method. Each configuration was only run once, but the results were verified through statistical significance tests that we explain in Section \ref{sec:data_and_eval}.

\begin{table}[thb]
\centering
\begin{tabular}{p{0.6\linewidth} | p{0.2\linewidth} }
\hline
Name & value\\
\hline
\# tokens in one gradient update & 524,288 \\
Learning rate & 0.0002 \\
Optimizer & Adafactor \\
Max sequence length & 128 \\
\hline
Expert dropout & 0.25 \\
Router hidden dimension & 384 \\
$\alpha$ & 0.1 \\
$\beta$ & 1.0 \\
\hline
\end{tabular}
\caption{\label{tab:hyper-param}
\# tokens in one gradient update is the maximum number of tokens in one batch $\times$ gradient accumulation $\times$ \# GPU.
}
\end{table}

\begin{table}[thb]
\centering
\begin{tabular}{p{0.6\linewidth} | p{0.3\linewidth} }
\hline
Name & value\\
\hline
\# tokens in one gradient update & \{524,288, 1,048,576\} \\
Learning rate & \{0.0002, 0.0004\} \\
Optimizer & Adafactor \\
Max sequence length & 128 \\
\hline
Expert dropout & \{0.25, 0.3\} \\
Router hidden dimension & 384 \\
$\alpha$ & \{0.1, 0.5\} \\
$\beta$ & \{0.1, 0.5, 1.0\} \\
\hline
\end{tabular}
\caption{\label{tab:hp-bound}
The hyper-parameter search bounds.
}
\end{table}

\end{document}